\documentclass[letterpaper]{article}
\usepackage{aaai19}
\usepackage{times}
\usepackage{helvet}
\usepackage{courier}
\usepackage{amsmath, amssymb, amsthm, mathtools}
\usepackage{color}
\usepackage{tabu}
\usepackage{ulem}

\newtheorem{theorem}{Theorem}
\newtheorem{result}{Result}
\newtheorem{prop}{Proposition}

\newcommand{\lw}[1]{\color{black}#1 \color{black}}

\begin{document}
\title{Improving Generalization of Sequence  Encoder-Decoder Networks for \\ Inverse Imaging of Cardiac Transmembrane Potential}
%\author{**}
\author{Sandesh Ghimire, Prashnna Kumar Gyawali, John L Sapp, Milan Horacek, Linwei Wang\\
Golisano College of Computing and Information Sciences\\
Rochester Institute of Technology\\
USA
}
\maketitle
\begin{abstract}
Deep learning models have shown state-of-the-art performance in many inverse reconstruction problems. However, it is not well understood what properties of the latent representation may improve the generalization ability of the network. Furthermore, limited models have been presented for inverse reconstructions over time sequences. 
In this paper, 
we study the generalization ability of 
a sequence encoder-decoder model 
for solving inverse reconstructions on time sequences. Our central hypothesis is that 
the generalization ability of the network 
can be improved by 1) constrained stochasticity and 2) global aggregation of temporal information 
in the latent space. 
First, drawing from analytical learning theory, we theoretically show that a stochastic latent space will lead to an improved generalization ability. 
Second, we consider an LSTM encoder-decoder architecture that compresses a  
global latent vector from 
all last-layer units in the LSTM encoder. 
This model is compared with alternative LSTM encoder-decoder architectures, each in deterministic and stochastic versions. 
%To measure the generalization ability 
%to different extent, 
%experiments are carried out on 
%test datasets carefully designed to be different from the training set at various levels.
The results demonstrate that the generalization ability of an inverse reconstruction network can be improved by constrained stochasticity combined with global aggregation of temporal information in the latent space. 

\end{abstract}
\section{Introduction}
{There has been an upsurge in the use of deep learning based methods in} 
inverse problems in computer vision and medical imaging \cite{lucas18}. {Examples include image} denoising \cite{mao16}, inpainting \cite{pathak2016context}, super resolution \cite{wang2015deep} and image reconstructions in \lw{a variety of medical imaging modalities such as} magnetic resonance imaging, X-ray, and computed tomography  \cite{jin17,wang2016accelerating,chen2017low}.  {A common deep network architecture for inverse reconstructions is a encoder-decoder network, which} {learns to encodes an input}  measurement into a latent representation that is then decoded into a desired %output image as the 
inverse solution \cite{pathak2016context,zhu18}. \lw{Despite significant successes of these models,} two important \lw{questions} remain \lw{relatively} unexplored. First, it is not well understood what properties of the latent representation improve the generalization ability of the network. 
Second, \lw{existing} works \lw{mostly} focus on solving single-image problems, \lw{while  limited models exist for solving inverse problems on image or signal sequences. The latter, however, is important because the incorporation of} \lw{temporal information can often help alleviate the ill-posedness of an inverse reconstruction problem.} 

{In this paper, we present a probabilistic sequence encoder-decoder model for solving inverse problems on time sequences.}
Central to \lw{the presented model is an emphasis on the} generalization ability \lw{of the network in order to} learn a general inverse \lw{mapping from the} measurement to reconstruction \lw{results.} \lw{Our main hypothesis is that} 
the generalization ability of the sequence encoder-decoder network can be improved by \lw{the following two properties of} the latent space: 1) \lw{constrained} stochasticity, \lw{and} 2) \lw{global aggregation of information throughout the long time sequence.} 

\lw{First,} we \lw{theoretically} show that using stochastic latent space during training helps to learn a decoder that is less sensitive to local changes in  the latent space and that, based on analytical learning theory \cite{kawaguchi18analytical},  
leads to good generalization ability \lw{of the network}. % based on analytical learning theory. 
\lw{Second, while it is common in a sequence model to use the last unit of a recurrent encoder network for decoding, we hypothesize that --} 
%\sout{The idea to use concise latent space stems from the motive to learn global feature of the whole sequence. The ECG at each instant is a weak projection of TMP signals on the body surface. However, the whole sequence of ECG could provide knowledge about the dynamics of wavefront propagation inside heart, which could be represented concisely. Based on this intuition, we propose to use Encoder-Decoder network with concise and summarizing latent representation. inspired by the works in sequence to sequence language translation \cite{sutskever14}, we} \sout{propose to use}  %\lw{will build on a foundation that consists of long-short-term-memory} (LSTM) networks in both the encoder and decoder to perform inverse reconstruction of signal/image sequences. 
%\lw{To learn a latent representation that summarizes the sequenc in this setting,} \lw{it is common in existing works to} 
%\sout{Using concise latent representation is not new; in the natural language processing (NLP) literature, there exists architectures that use LSTM in the encoder and decoder and vector as latent space, for example language translation models that} 
%use \lw{the} last \lw{unit}\sout{hidden layer} of \lw{the} LSTM network 
%as the \lw{the}
% \sout{concise} representation \lw{that will be used for} \sout{and} reconstruction \sout{the output sequence from that single vector} \cite{sutskever14}. However,
\lw{in the presence of long time sequences --} the last hidden unit code  
%\sout{have some limitations: 1) the last hidden code might not be rich in information and act as a bottleneck in information flow, 2) for long sequences, like in our problems, such architectures cannot}
\lw{may not be able to} retain \lw{global information and  may act as a bottleneck in information flow.}  
%In addition, the issue of gradient vanishing may arise as the length of the sequence grows. These issues have been echoed in recent works, } \sout{rich information and exhibit problems during back propagation. Recent papers have expressed similar concerns and proposed}  
\lw{This shares a fundamental idea with recent works in which alternative} architectures \lw{were presented} with attention mechanism \lw{to} consider  information from all the hidden states of long-short-term-memory (LSTM) networks  \cite{bahdanau14,luong15}. \lw{Alternative strategies were also presented to} back-propagate \lw{the} loss \lw{through} all hidden states {addressing the issue of gradient vanishing} \cite{lipton15,yue15}.  \lw{Here, to arrive at a compact latent space that globally summarizes a long time sequence,} 
%\sout{our case, on the one hand we need a summarizing vector as latent representation to capture the essence of sequence into a vector; on the other hand, we also want this representation to be rich and summarizing long sequence . To achieve this contradictory requirement,}
we \lw{present an architecture (termed as \textit{svs}) to combine and compress} all LSTM \lw{units into a vector representation.} 
%\lw{through fully-connected layers}
%and further compress them into single latent vector 
%(see Fig.\ref{architecture}). 

While the presented methodology applies for general sequence inverse problems, 
we test it on \lw{the problem of} inverse reconstruction of cardiac transmembrane potential (TMP) sequence from {high-density electrocardiogram} (ECG) sequence {acquired on the body surface 
%This problem has important clinical applications such as in development of diagnostic and therapeutic tools for ischemia, ventricular arrhythmias and infarction 
\cite{macleod95,ramanathan2004,wang13}. 
In this setting, the measurement at each time instant is a spatial potential map on the body surface, and the reconstruction output is a spatial potential map throughout the three-dimensional heart muscle. The problem at each time instant is severely ill-posed, and incorporating temporal information is recognized as a main approach to alleviate this issue.}   
%Inverse imaging of TMP provides a non-invasive alternative to access electrical activity of the heart and has important clinical applications such as in development of diagnostic and therapeutic tools for ischemia \cite{macleod95}, ventricular arrhythmias \cite{ramanathan2004} and infarction \cite{wang13}. 

To investigate {the} generalization ability of {the presented model, we 
analyzed in-depth the benefits brought by each of the two key components within the presented model. In specific, we compared the presented \textit{svs} architecture with two alternatives: one that decodes directly from the sequence produced by the LSTM encoder (termed as \textit{sss}) and one that decodes from the output of the last unit of the LSTM encoder, as mostly commonly used in language models (termed as \textit{svs-L}) \cite{sutskever14}. We also compared between deterministic and stochastic versions of \textit{svs} and \textit{svs-L}. %Comparison was carried out on} test datasets \lw{carefully designed to be} \sout{with data} different from \lw{the} training set at different levels, \lw{allowing us to comprehensively} \sout{which} measure \lw{the extent to which each of the network is able to} generalize. 
\lw{The experiments results suggest that the generalization ability of the network in inverse reconstruction can be improved by constrained stochasticity combined with global aggregation of sequence information in the latent space. These findings may set} %\sout{By comparing performance of stochastic latent space against deterministic and sequential latent space against vector, we understand role of different design choices of latent space for improving generalization ability of networks. we believe this work will lay}
a foundation for investigating the generalization ability of \lw{deep networks in} sequence inverse problems and inverse problems in general.

\section{Related Work}
There is a large body of work in the \lw{the use of deep learning in} inverse imaging in 
both general computer vision \cite{lucas18,mao16,wang2015deep,yao2017dr2,fischer2015flownet} and medical imaging \cite{jin17,wang2016accelerating,chen2017low}. 
\lw{Some of these deep inverse reconstruction networks are 
based on an encoder-decoder structure \cite{pathak2016context,zhu18}, similar to that 
investigated in this paper. 
Among these works in}
 the domain of medical image reconstruction, 
 \lw{the presented}
 work is the closest to Automap \cite{zhu18}, \lw{in that} the output image is reconstructed directly from the \lw{input} measurements without \lw{any} domain-specific  intermediate transformations. 
 However, all \lw{of} these works employ a deterministic architecture \lw{which,} as we will show later, can be improved with \lw{the introduction of} stochasticity. 
 \lw{Nor do these existing works  
 handle inverse reconstructions of images or signals over time sequences.}

Different elements of \lw{the presented} work are conceptually similar to several works across different domains of machine learning. \lw{The use of encoder-decoder architectures for sequential signals is related to existing}
works in language translation 
%are also related to our work in the use of Encoder-Decoder architecture for sequential signals, and 
%using concise latent space representation 
\cite{sutskever14}. However, 
\lw{we investigate an alternative 
global aggregation of sequence information 
by utilizing and compressing}
knowledge from all the \lw{units in the last} layer of 
the LSTM encoder. 
\lw{This is in concept similar to the works in}
%Similarly, 
\cite{bahdanau14,luong15}, 
\lw{where} information 
from all \lw{the units of an LSTM encoder are used} for language translation. 
\lw{However, 
to our knowledge, 
no existing works have analyzed in-depth 
the difference in 
the generalization ability of sequence encoder-decoder models 
with respect to these different designs of the latent space.  
}

The presented theoretical analysis of 
stochasticity in generalization utilizes analytical learning theory \cite{kawaguchi18analytical}, which is fundamentally different from classical statistical learning theory in that it is strongly instance-dependent. While statistical learning theory deals with data-independent generalization bounds or data dependent bounds for certain hypothesis space of problems, analytical learning theory 
%provides generalization bounds for a specific problem instance: it 
provides the bound on how well a model learned from a dataset should perform on true (unknown) measures of variable of interest. This makes it aptly suitable for measuring the generalization ability of a stochastic latent space for the given problem and data.

The presented work is related to variational autoencoder (VAE) in using stochastic latent space with regularization \cite{kingma13}. Similarly, \cite{bowman15} present a sequence-to-sequence VAE based on LSTM encoder and decoders to generate coherent and diverse sentences from continuous sampling of latent code. However, it is not well understood why stochasticity of the latent space is so important. In this paper, we intend to provide a justification from the learning theory perspective. \lw{In addition, while VAE by nature is concerned with the reconstruction of the same input data, the presented network is concerned with the ill-posed problem of inverse signal/image reconstruction from their (often weak) measurements.}
%is different from VAE since we do not reconstruct the same input signal as output.

%\lw{[how strong do you feel about including this?]} Context encoders \cite{pathak2016context}, for example, generate contents of a region in image based on its surroundings. If we think of whole sequence as providing global temporal context in our case, then we may be able to draw some connection between our work and Context encoders. Like ours, Context encoder also uses Encoder-Decoder architecture, but with convolutional and fully connected layers as the context is spatial rather than temporal in their case.

\section{Preliminaries}
\subsection{Inverse Imaging of  {cardiac transmembrane potential} (TMP)} 
\lw{Body-surface electrical potential is produced by TMP in the heart. Their} mathematical relation 
%between cardiac TMP 
%and the body-surface potential produced by them 
%ECG signals produced by them on body surface can be obtained 
is defined by the quasi-static approximation of electromagnetic theory 
{\cite{plonsey1969bioelectric} and, when solved on patient-specific heart-torso geometry, can be derived as}  
%Using bidomain theory and maxwell equations, we can reduce the biophysical relation between the ECG, $\boldsymbol{y}$ and TMP, $\boldsymbol{x}$ by the relation 
\cite{wang10}: 
\begin{align}
\boldsymbol{y}=\boldsymbol{Hx}
\end{align}
where $\boldsymbol{y}$ denotes the body-surface potential map, $\boldsymbol{x}$ the 3D TMP map over the heart muscle, % \cite{wang10}
and $\boldsymbol{H}$ the measurement matrix specific to the heart-torso geometry of an individual. 

\lw{The inverse reconstruction of $\boldsymbol{x}$ from $\boldsymbol{y}$ 
can be carried out at each time instant independently, which however is notoriously ill-posed since surface potential provides only a weak projection of the 3D TMP. A popular approach is thus to reconstruct the time sequence of TMP propagation over a heart beat, with various strategies to incorporate the temporal information to alleviate the ill-posedness of the problem \cite{wang10,greensite98}. Here, we examine the sequence setting, where 
$\boldsymbol{x}$ and $\boldsymbol{y}$ represents 
sequence matrices with each column 
denoting the potential map at one time instant.} This problem has important clinical applications 
in supporting the diagnosis and treatment for 
diseases such as 
ischemia \cite{macleod95}  
%infarction \cite{wang13}, 
and ventricular arrhythmia \cite{Wang_Europace18}.

%and both $\boldsymbol{y}$ and $\boldsymbol{x}$ are matrices with each column denoting signals at a time instant. The inverse problem then becomes estimation of $\boldsymbol{x}$ given $\boldsymbol{y}$. 
\begin{figure*}[t!]
\centering
\includegraphics[width=0.7\linewidth]{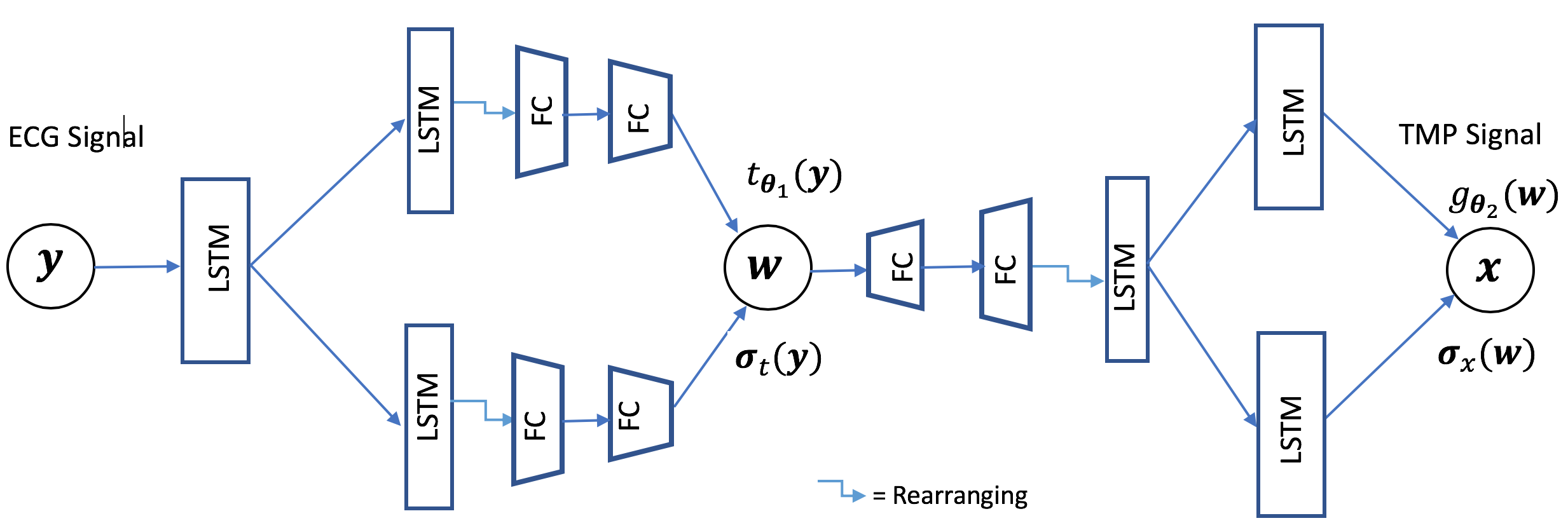}
\caption{\small{Presented \textit{svs stochastic} architecture with mean and variance network in both encoder and decoder.}}
\label{architecture}
\end{figure*}
\subsection{RNN and LSTM}
Recurrent neural networks (RNN) \cite{werbos1990backpropagation,kalchbrenner2013recurrent} generalize the traditional neural networks for sequential data by allowing passage of information over time. 
{Compared to traditional RNNs, 
LSTM networks \cite{graves2013generating,sutskever14} can better handle long term dependency in the data 
by architectural changes such as} 
using a memory line called the cell state that 
runs throughout the whole sequence, 
%\sout{However, RNNs become ineffective (due to reasons like vanishing gradient) when long term dependency needs to be modeled in data. 
%Long Short-Term Memory 
%LSTM networks were developed to address these limitations of RNNs}   .  
%in which 
the ability to forget information deemed irrelevant using a forget gate, 
and the ability to selectively update the cell state. 
%\sout{Because of these changes, LSTMs have been successful in modeling long range dependencies.}\lw{[I know why this is here, but it's creating conflicting messages with earlier motivations. We're not reviewing here but to provide background; may be sufficient to keep it short]}

\subsection{Variational Autoencoder (VAE)} 
VAE \cite{kingma13} is a \lw{probabilistic} generative model \lw{that is typically trained} by optimizing \lw{the} variational lower bound of \lw{the} data log likelihood. A VAE is distinctive from the traditional autoencoder \lw{in two aspects}: 1) \lw{the use of} stochastic latent space realized by sampling with a reparameterization trick, 2) \lw{the use of} Kullback-Leibler (KL) divergence to regularize the latent distribution. 

%\lw{I'm not sure yet, but how do you feel about talking about the stochastic rationale earlier in the paper, before the svs architecture?, i.e., give the overall LSTM and VAE setup first, then talk about the stochastic rationale, then talk about the architecture in terms of svs, etc. The only concern is that the stochastic part may be too long and bury the second part....Or, areyou able to give the theoretical stochastic analysis first to set the stage, before starting to talk about your model at all -- may question is if you can talk about your results 1 before giving the VAE setup...Also, depending on the order you decide to talk about these two things in method, that should be the same order they're motivated in intro as well -- usually you want to give the primary contribution first.}
\section{Methodology}
We train a probabilistic sequence encoder-decoder network  
to learn to reconstruct the time sequence of TMP, $\boldsymbol{x}$, 
from input body-surface potential, $\boldsymbol{y}$. 
In a supervised setting, 
we  
maximize the log likelihood %in a supervised setting 
as follows: 
\begin{align}
\underset{\boldsymbol{\theta}}{argmax} \hspace{0.2cm}{E_{P(\boldsymbol{x},\boldsymbol{y})}\log p_{\boldsymbol{\theta}}(\boldsymbol{x}|\boldsymbol{y})}
\end{align}
where $P(\boldsymbol{x},\boldsymbol{y})$ is the joint distribution of the input-output pair. We introduce a latent random variable $\boldsymbol{w}$ and express the conditional distribution as:  
\begin{align}
 p_{\boldsymbol{\theta}}(\boldsymbol{x}|\boldsymbol{y})=\int p_{\boldsymbol{\theta}_2}(\boldsymbol{x}|\boldsymbol{w})p_{\boldsymbol{\theta}_1}(\boldsymbol{w}|\boldsymbol{y})d\boldsymbol{w}
\end{align}
where $\boldsymbol{\theta}=\{\boldsymbol{\theta}_1, \boldsymbol{\theta}_2 \}$. We model both $p_{\boldsymbol{\theta}_2}(\boldsymbol{x}|\boldsymbol{w})$ and $p_{\boldsymbol{\theta}_1}(\boldsymbol{w}|\boldsymbol{y})$ 
with Gaussian distributions, 
with mean and variance % of both the distributions are 
parameterized by neural networks: 
\begin{align}
\label{encoder}
      p_{\boldsymbol{\theta}_1}(\boldsymbol{w}|\boldsymbol{y})=\mathcal{N}(\boldsymbol{w}|\boldsymbol{t}_{\theta_1}(y),\boldsymbol{\sigma_t}^2(\boldsymbol{y}))\\
\label{decoder}
  p_{\boldsymbol{\theta}_2}(\boldsymbol{x}|\boldsymbol{w})=\mathcal{N}(\boldsymbol{x}|\boldsymbol{g}_{\theta_2}(\boldsymbol{w}),\boldsymbol{\sigma_x}^2(\boldsymbol{w}))
\end{align}
where $\boldsymbol{\sigma_x}^2$ denotes a matrix of the same dimension as that of $\boldsymbol{x}$. We implicitly assume  that each elements in $\boldsymbol{x}$ is independent and Gaussian with variance given by the corresponding element in $\boldsymbol{\sigma_x}^2$; and similarly for $\boldsymbol{w}$. %Also note abuse of notation $\mathcal{N}$ in eq.(\ref{encoder}, \ref{decoder}).
Introduction of the latent random variable 
in the network 
allows us to constrain it 
in two means  
to improve the generalization ability 
of inverse reconstructions. 
First, 
we constrain the conditional distribution $p_{\boldsymbol{\theta}_1}(\boldsymbol{w}|\boldsymbol{y})$ to be close to an isotropic Gaussian distribution. 
Second, we design it to be a 
concise vector representation 
compressed from the whole input time sequence.
%\subsection{Constraining latent space}

\subsection{Regularized stochasticity }% of the latent space}

Drawing from the VAE \cite{kingma13}, 
we regularize the latent space 
by constraining the 
conditional distribution $p_{\boldsymbol{\theta}_1}(\boldsymbol{w}|\boldsymbol{y})$ to be close to 
an isotropic Gaussian distribution. 
Training of the network 
can then be formulated as 
a constrained optimization problem as follows:
\begin{align}
\nonumber
&\textrm{minimize} -{E_{P(x,y)}\log \int p_{\boldsymbol{\theta}_2}}(\boldsymbol{x}|\boldsymbol{w})p_{\boldsymbol{\theta}_1}(\boldsymbol{w}|\boldsymbol{y})d\boldsymbol{w}\\
&\textrm{such} \hspace{0.1cm}\textrm{that} \hspace{0.1cm} D_{KL}(p_{\boldsymbol{\theta}_1}(\boldsymbol{w}|\boldsymbol{y})||\mathcal{N}(\boldsymbol{w}|\boldsymbol{0},\boldsymbol{I}))<\delta 
\end{align}
Using the method of Lagrange multipliers, we reformulate the objective function into:
\begin{align}
\nonumber
\mathcal{L}&= -{E_{P(x,y)}\log \Big[ \int p_{\boldsymbol{\theta}_2}}(\boldsymbol{x}|\boldsymbol{w})p_{\boldsymbol{\theta}_1}(\boldsymbol{w}|\boldsymbol{y})d\boldsymbol{w}\\
\nonumber
&+\lambda.(D_{KL}(p_{\boldsymbol{\theta}_1}(\boldsymbol{w}|\boldsymbol{y})||\mathcal{N}(\boldsymbol{w}|\boldsymbol{0},\boldsymbol{I}))-\delta) \Big]\\
\nonumber
\label{lowerbound}
&\leq -{E_{P(x,y)} \Big[ E_{w\sim p_{\boldsymbol{\theta}_1}(\boldsymbol{w}|\boldsymbol{y})\}}[ \log p_{\boldsymbol{\theta}_2}}(\boldsymbol{x}|\boldsymbol{w})]\\
&+\lambda.(D_{KL}(p_{\boldsymbol{\theta}_1}(\boldsymbol{w}|\boldsymbol{y})||\mathcal{N}(\boldsymbol{w}|\boldsymbol{0},\boldsymbol{I}))-\delta) \Big]
\end{align}
where the inequality in eq.(\ref{lowerbound}) is due to Jensen's inequality 
as the negative logarithm is a convex function. 
We use reparameterization $\boldsymbol{w}=\boldsymbol{t}+\boldsymbol{\sigma}_t\odot \boldsymbol{\epsilon}$ as described in \cite{kingma13} to compute 
the inner expectation in the first term. 
The KL divergence in the second term 
is analytically available for two Gaussian distributions. 
We thus obtain the upper bound for the loss function as: 
\begin{align}
\nonumber
&\mathcal{L}\leq E_{P(\boldsymbol{x},\boldsymbol{y})} \Big[ E_{\boldsymbol{\epsilon}\sim\mathcal{N}(\boldsymbol{0},\boldsymbol{I})}\Big( \sum_i\frac{1}{\boldsymbol{\sigma_x}_{i}^2}(x_i-g_i(\boldsymbol{t}+\boldsymbol{\sigma_t}\odot \boldsymbol{\epsilon}))^2\\
\label{loss}
&+\log \boldsymbol{\sigma_x}_{i}^2\Big)
+\lambda . D_{KL}(p_{\boldsymbol{\theta}_1}(\boldsymbol{w}|\boldsymbol{y})||\mathcal{N}(\boldsymbol{w}|\boldsymbol{0},\boldsymbol{I})) \Big]+constant
\end{align}
where $g_i$ is the $i^{th}$ function mapping latent variable to the  $i^{th}$ element of mean of $\boldsymbol{x}$, such that $\boldsymbol{g}_{\boldsymbol{\theta}_2}=[g_1,g_2...g_U]$.
%\subsection{Latent dropout}

\subsection{Global aggregation of sequence information} 
In sequence inverse problems %like TMP imaging 
where the measurement at each time instant 
provides only a weak projection of the reconstruction solution, 
utilizing the temporal information in the sequence 
becomes important for better inverse reconstructions. 
This motivate us to 
design an architecture 
that can distill from the input sequence a global, 
%The hypothesize that using concise latent space helps generalization is backed by the domain knowledge in inverse imaging of ECG.
%The cardiac transmembrane potential follows certain dynamics producing a wavefront of electrical propagation in the heart. Therefore, TMP propagation could be succinctly expressed by a 
time-invariant, and low dimensional 
latent representation 
from which the entire TMP sequence can be reconstructed. 

To do so, 
we present an architecture with two LSTM networks followed by two fully connected neural networks (FC), each respectively for the mean and variance in the encoder network. The decoder then consists of two FC followed by two LSTM networks for the mean and variance of the output. In the encoder, each LSTM decreases spatial dimensions while keeping temporal dimensions constant; the last-layer outputs from all the units in each LSTM are reshaped into a vector, the length of which is decreased by the FC. The structure and dimension of the decoder mirrors that of the encoder.  
%, the dimension is increased correspondingly.  
%before applying FCNN in the encoder and the vector from FCNN is restructured into matrix in the same order before applying LSTM in the decoder. 
The overall architecture of the presented network is illustrated in Fig.~\ref{architecture}. 

%. On the other hand, ECG sequence carries information about the whole TMP dynamics even if ECG at individual time stamp might be weak projection of TMP. Based on this intuition, using an encoder-decoder architecture with concise latent vector and reconstructing whole sequence of TMP from it helps the network learn a concise, global representation, which later helps in generalization.

\section{Encoder-Decoder Learning from the Perspective of Analytical Learning Theory}
In this section we look at the encoder-decoder inverse reconstructions from the 
analytical learning theory \cite{kawaguchi18analytical}. We start with a deterministic latent space setting and then show that having a stochastic latent space with regularization helps in generalization.

Let $\boldsymbol{z}=(\boldsymbol{y},\boldsymbol{x})$ be an input-output pair, and let $D_n=\{\boldsymbol{z}^{(1)},\boldsymbol{z}^{(2)},...,\boldsymbol{z}^{(n)}\}$ denote the total set of training and validation data and $Z_m \subset D_n$ be a validation set. During training, a neural network learns the parameter $\boldsymbol{\theta}$ by using an algorithm $\mathcal{A}$ and dataset $D_n$, at the end of which we have a mapping $h_{\mathcal{A}(D_n)}(.)$ from $\boldsymbol{y}$ to $\boldsymbol{x}$.
%At the end of training, we have mean and variance functions of $x$ as $\bar{x}_{\mathcal{A}(D_m)}(y)$ and $\boldsymbol{\sigma}_{x\mathcal{A}(D_m)}(y)$ respectively. We define a loss function based on our notion of goodness of prediction as $\ell(x,\bar{x}_{\mathcal{A}(D_m)}(y), \boldsymbol{\sigma}_{x\mathcal{A}(D_m)}(y))$. NOte that $\ell$ could be mean square loss or negative log likelihood and is not required to be the same as that used in training. The average validation error is given by ${E_{Z_m}l(x,g_{\theta(A(D_m))}(y))}$. 
Typically, we stop training when the model performs well in the validation set. To evaluate this performance, we define a loss function based on our notion of goodness of prediction as $\ell(\boldsymbol{x},h_{\mathcal{A}(D_n)}(\boldsymbol{y}))$. The average validation error is given by ${E_{Z_m}\ell(\boldsymbol{x},h_{\mathcal{A}(D_n)}(\boldsymbol{y}))}$. 

%The validation error gives us a  sense of how well the model performs during the test. 
However, there exists a gap between how well the model performs in the validation set versus in the true distribution of the input-output pair; this gap is called the generalization gap. To be precise let $(\mathcal{Z},\mathcal{S},\mu)$ be a measure space with $\mu$ being a measure on $(\mathcal{Z},\mathcal{S})$. Here, $\mathcal{Z}=\mathcal{Y}\times\mathcal{X}$ denotes the input-output space of all the observations and inverse solutions.  The generalization gap is given by:
\begin{align}
\label{eqn:gap}
{E_{\mu}\ell(\boldsymbol{x},h_{\mathcal{A}(D_n)}(\boldsymbol{y}))}-{E_{Z_m}\ell(\boldsymbol{x},h_{\mathcal{A}(D_n)}(\boldsymbol{y}))}
\end{align}
Note that this generalization gap depends on the specific problem instance. Theorem 1 \cite{kawaguchi18analytical}  provides an upper bound on equation (\ref{eqn:gap}) 
%this generalization gap 
in terms of data distribution in the latent space and properties of the decoder.
\begin{theorem}[\cite{kawaguchi18analytical}]
For any $\ell$, let $(\mathcal{T},f)$be a pair such that $\mathcal{T}:(\mathcal{Z},\mathcal{S})\rightarrow ([0,1]^d,\mathcal{B}([0,1]^d))$ is a measurable function, $f:([0,1]^d,\mathcal{B}([0,1]^d))\rightarrow (\mathbb{R},\mathcal{B}(\mathbb{R}))$ is of bounded variation as $V[f]<\infty$, and
$\ell(\boldsymbol{x}, h(\boldsymbol{y}))=(f\circ \mathcal{T})(\boldsymbol{z}) \forall \boldsymbol{z} \in \mathcal{Z}$, where $\mathcal{B}(A)$ indicates the Borel $\sigma$- algebra on $A$. Then for any dataset pair $(D_m,Z_m)$ and any $\ell(\boldsymbol{x},h_{\mathcal{A}(D_n)}(\boldsymbol{y}))$,
\begin{align*}
{E_{\mu}\ell(\boldsymbol{x},h_{\mathcal{A}(D_m)}(\boldsymbol{y}))}-{E_{Z_m}\ell(\boldsymbol{x},h_{\mathcal{A}(D_n)}(\boldsymbol{y}))}\\
\leq V[f]\mathcal{D}^*[\mathcal{T}_*\mu, \mathcal{T}(Z_m)]
\end{align*}
where $\mathcal{T}_*\mu$ is pushforward measure of $\mu$ under the map $\mathcal{T}$.
\end{theorem}

For an encoder-decoder setup, $\mathcal{T}$ is the encoder which maps the observation to the latent space and $f$ becomes the composition of loss function and decoder which maps latent representation to the reconstruction loss. Note that the  $[0,1]^d$ latent domain can be easily extended to a d-orthotope -- as long as the latent variables are bounded -- using a function $s$ composed of scaling and translation in each dimension. Since $s$ is uniformity preserving and affects the partial derivative of $f$ only up to a scaling factor and thus does not affect our analysis.  In practice, there always exists intervals such that the latent representations are bounded. 

Theorem 1 provides two ways to decrease the generalization gap in our problem: by decreasing the variation $V[f]$ or the  discrepancy $\mathcal{D}^*[\mathcal{T}_*\mu, \mathcal{T}(Z_m)]$. Here, we show that constrained stochasticity of the latent space helps decrease the variation $V[f]$. 
%, thereby improving generalization. 
The variation of $f$ on $[0,1]^d$ in the sense of Hardy and Krause \cite{hardy}  is defined as: 
\begin{align}
V[f]=\sum_{k=1}^d\sum_{1\leq j_1<...<j_k\leq d}V^k[f_{j_1...j_k}]
\end{align}
where $V^k[f_{j_1...j_k}]$ is defined with following proposition.
\begin{prop}[\cite{kawaguchi18analytical}]
Suppose  that $f_{j_1,..j_k}$ is a function for which $\partial_{1,...k} f_{j_1,..j_k}$ exists on $[0,1]^k$. Then,\\
$V^k[f_{j_1...j_k}]\leq \underset{\boldsymbol{t}_{j_1},..,\boldsymbol{t}_{j_k}\in [0,1]^k}{sup}|\partial_{1,...k} f_{j_1,..j_k}(\boldsymbol{t}_{j_1},..,\boldsymbol{t}_{j_k})|$.\\
If $\partial_{1,...k} f_{j_1,..j_k}$ is also continuous on $[0,1]^k$,
$$V^k[f_{j_1...j_k}]=\int_{[0,1]^k}|\partial_{1,...k} f_{j_1,..j_k}(\boldsymbol{t}_{j_1},..,\boldsymbol{t}_{j_k})|dt_{j_1}..dt_{j_k}.$$
\end{prop}

The function $f$ for the encoder decoder network is the loss $\ell$ as a function of latent representations. Thus, we have,
\begin{align}
\partial_{1,...k} f_{j_1,..j_k}(\boldsymbol{t}_{j_1},..,\boldsymbol{t}_{j_k})=\frac{\partial ^k \ell}{\partial \boldsymbol{t}_{j_1},..,\partial \boldsymbol{t}_{j_k}}
\end{align}
We use a simple sum of square loss:
%\begin{align}
$
\ell(\boldsymbol{x}, h(\boldsymbol{y}))=||\boldsymbol{x}-\boldsymbol{g}_{\boldsymbol{\theta}_2}(\boldsymbol{t})||^2=\sum_i (\boldsymbol{x}_i-g_i(\boldsymbol{t}))^2
$
%\end{align}
where the norm is frobenius norm for matrix $\boldsymbol{x}$, and $g$ is a function from latent space to each element of estimated $\bar{\boldsymbol{x}}$. Writing $\ell_i=(\boldsymbol{x}_i-g_i(\boldsymbol{t}))^2$, 
\begin{align}\label{sumloss}
\ell(\boldsymbol{x}, h(\boldsymbol{y}))=\sum_i\ell_i
\end{align}

Theorem 1 and Proposition 1 implies that if the  cross partial derivative of loss with respect to the latent vector at all order is low in all directions throughout the latent space, then the approximated validation loss would be closer to the actual loss over the true unknown distribution of the dataset. Intuitively, we want the loss curve as a function of latent representation to be flat if we want a good generalization. 
\subsection{Using stochastic latent space}
In our formulation, the latent space is a random variable with the cost function given by eq.(\ref{loss}), which makes the latent vector stochastic by design. The inner expectation of first term in the cost function is given by
\begin{align}
\nonumber
T_1&=E_{\boldsymbol{\epsilon}\sim\mathcal{N}(\boldsymbol{0},\boldsymbol{I})}[\sum_i\frac{1}{\boldsymbol{\sigma}_{xi}^2}(\boldsymbol{x}_i-g_i(\boldsymbol{t}+\boldsymbol{\sigma}_t \odot \boldsymbol{\epsilon}))^2]\\
%\label{sumofsquare}
\nonumber
&=\sum_i \frac{1}{\boldsymbol{\sigma}_{x i}^2}(\boldsymbol{x}_i-g_i(\boldsymbol{t}+\boldsymbol{\eta}))^2%\\
=\sum_i \frac{1}{\boldsymbol{\sigma}_{x i}^2} E_{\boldsymbol{\epsilon}}[\ell_i(\boldsymbol{x}_i,\boldsymbol{t}+\boldsymbol{\eta})]
\end{align}
where $\boldsymbol{\eta}=\boldsymbol{\sigma}_t\odot\boldsymbol{\epsilon}$. 
\begin{result}
\begin{align*}
T_1&=\sum_i  \frac{1}{\boldsymbol{\sigma}_{x i}^2} \Big[ \ell_i(\boldsymbol{x}_i,\boldsymbol{t})+\langle \boldsymbol{\sigma}_t \odot E_{\epsilon}[\boldsymbol{\epsilon}],\frac{\partial}{\partial t}\ell_i(\boldsymbol{x}_i,\boldsymbol{t})\rangle\\
&
+\frac{1}{2}\langle[\boldsymbol{\sigma}_t\otimes\boldsymbol{\sigma}_t]\odot E_{\epsilon}[\boldsymbol{\epsilon}\otimes\boldsymbol{\epsilon}], \Big[ \frac{\partial ^2}{\partial \boldsymbol{t}_{j_1},\partial \boldsymbol{t}_{j_2}}\ell_i(\boldsymbol{x}_i,\boldsymbol{t}) \Big]\rangle+..\\
&+\frac{1}{k!}\langle[\boldsymbol{\sigma}_t\otimes^k\boldsymbol{\sigma}_t]\odot E_{\epsilon}[\boldsymbol{\epsilon}\otimes^k\boldsymbol{\epsilon}], \Big[ \frac{\partial ^k}{\partial \boldsymbol{t}_{j_1},..,\partial \boldsymbol{t}_{j_k}}\ell_i(\boldsymbol{x}_i,\boldsymbol{t})\Big]\rangle+.. \Big]
\end{align*}
where $[\boldsymbol{\sigma}_t\otimes^k\boldsymbol{\sigma}_t]$ denotes k order tensor product of a vector $\boldsymbol{\sigma}_t$ by itself.
\end{result} 
%Proof of Result 1 is provided in the supplemental material. 
\begin{proof}
 Using Taylor series expansion for $\ell_i(\boldsymbol{x}_i,\boldsymbol{t}+\boldsymbol{\eta})$,
 \begin{align}
 \nonumber
 &E_{\epsilon}[\ell_i(\boldsymbol{x}_i,\boldsymbol{t}+\boldsymbol{\eta})]=E_{\epsilon}\Big[\ell_i(\boldsymbol{x}_i,\boldsymbol{t})+\langle\boldsymbol{\eta},\frac{\partial}{\partial \boldsymbol{t}}\ell_i(\boldsymbol{x}_i,\boldsymbol{t})\rangle\\
 \nonumber
 &
 +\frac{1}{2}\langle[\boldsymbol{\eta}\otimes\boldsymbol{\eta}],\Big[ \frac{\partial ^2}{\partial \boldsymbol{t}_{j_1},\partial \boldsymbol{t}_{j_2}}\ell_i(\boldsymbol{x}_i,\boldsymbol{t}) \Big]\rangle+..\\
 \label{taylor}
 &+\frac{1}{k!}\langle[\boldsymbol{\eta}\otimes^k\boldsymbol{\eta}], \Big[ \frac{\partial ^k}{\partial \boldsymbol{t}_{j_1},..,\partial \boldsymbol{t}_{j_k}}\ell_i(\boldsymbol{x}_i,\boldsymbol{t})\Big]\rangle+..\Big]
 \end{align}
We move expectation operator inside both brackets and take expectation of only the first term in the inner product. Using $\boldsymbol{\eta}=\boldsymbol{\sigma}_t\odot\boldsymbol{\epsilon}$, we get $E_{\epsilon}[\boldsymbol{\eta}\otimes^k\boldsymbol{\eta}]=[\boldsymbol{\sigma}_t\otimes^k\boldsymbol{\sigma}_t]\odot E_{\epsilon}[\boldsymbol{\epsilon}\otimes^k\boldsymbol{\epsilon}]$. Using these in eq.(\ref{taylor}) yields the required result.
\end{proof}
%Below we examine implications of Result 1 as follows:
%\begin{enumerate}
%\item 
The first term of Result 1, $\ell_i(\boldsymbol{x}_i,\boldsymbol{t})$ (after ignoring $\frac{1}{\boldsymbol{\sigma}_{x i}^2}$), would be the only term in the cost function if the latent space were deterministic. Thus, the rest of the terms in Result 1 are additional in stochastic training. 
%Moreover, $T_1$ is non-negative from eq.(\ref{sumofsquare}).
%\item 
Each of these terms is an inner product of two tensor, the first being  $[\boldsymbol{\sigma}_t\otimes^k\boldsymbol{\sigma}_t]\odot E_{\epsilon}[\boldsymbol{\epsilon}\otimes^k\boldsymbol{\epsilon}]$, and the second being the $k^{th}$ order partial derivative tensor $\Big[ \frac{\partial ^k}{\partial \boldsymbol{t}_{j_1},..,\partial \boldsymbol{t}_{j_k}}\ell_i(\boldsymbol{x}_i,\boldsymbol{t})\Big]$.  We can thus consider the first tensor as providing penalizing weights to different partial derivatives in the second tensor. Since each inner product is added to the cost, we are minimizing them during optimization. This gives two important implications: 

\begin{enumerate}
\item For sufficiently large samples, $E_{\epsilon}[\boldsymbol{\epsilon}\otimes^k\boldsymbol{\epsilon}]$ must be close to central moments of isotropic Gaussian. However, in practice, the number of samples remains constant, but the number of parameters to be estimated keeps increasing for higher order tensors and reaches the order of the number of samples pretty quickly (forth order in our case). When the number of parameters is high, we can expect that those higher moments do not converge to that of standard Gaussian. This, luckily, works in our favor. Since we are minimizing $\frac{1}{k!}\langle[\boldsymbol{\sigma}_t\otimes^k\boldsymbol{\sigma}_t]\odot E_{\epsilon}[\boldsymbol{\epsilon}\otimes^k\boldsymbol{\epsilon}], \Big[ \frac{\partial ^k}{\partial \boldsymbol{t}_{j_1},..,\partial \boldsymbol{t}_{j_k}}\ell_i(\boldsymbol{x}_i,\boldsymbol{t})\Big]\rangle$ for each order, the inner product can be vanished for arbitrary $\epsilon$ only by driving partial derivative tensors towards zero. %Due to this minimization and relation of each tensor to its higher order tensor, 
Therefore, 
minimizing the sum of all the inner product for arbitrary $\epsilon$ would minimize most of the terms in the partial derivative tensor. From Proposition 1, minimizing each of these partial derivatives corresponds to minimization of variation of of function $\ell_i$, and consequently variation of the total loss $\ell$  according to eq.(\ref{sumloss}). Hence, additional terms in the stochastic latent space formulation contributes in decreasing variation of the loss.

\item Not all the partial derivatives are equally weighted in the cost function. Due to the presence of weighting tensor $[\boldsymbol{\sigma}_t\otimes^k\boldsymbol{\sigma}_t]$ in the first tensor of inner product, different partial derivative terms are penalized differently according to the value of $\boldsymbol{\sigma}_t$. 
%The KL divergence term in eq.(7?) encourages variance to be around 1, which is high value of variance. 
Combination of the KL divergence term in eq.(\ref{loss}) 
with $T_1$ thus tries to increase variance towards 1 
whenever it does not significantly increase the cost $T_1$:  higher value of $\boldsymbol{\sigma}_t$ %in certain direction 
penalizes the partial derivatives of a certain direction more heavily, % and encourage them to stay low, 
making the cost flatter in some directions than other.
\end{enumerate}

Strictly speaking, Proposition 1 requires cross partial derivatives to be small throughout the domain of latent variable, which is not included in the above analysis. 
It however should not significantly affect the observation 
that, compared to deterministic formulation, the stochastic formulation decreases the variation $V[f]$.

\begin{figure*}[t!]
\centering
\includegraphics[width=0.8\linewidth]{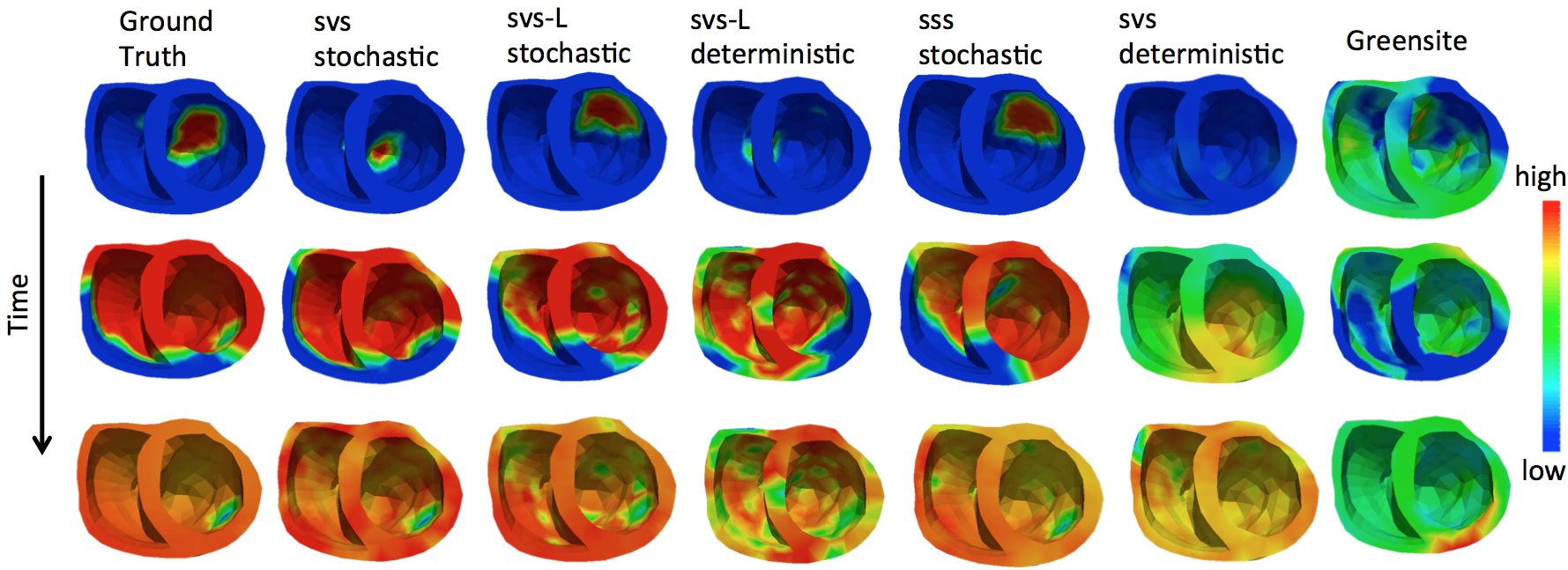}
\caption{\small{Comparison of TMP propagation over heart reconstructed from same ECG using different architecture. The propagation sequence is closer to ground truth in case of TMP reconstruction with svs stochastic architecture}}
\label{tmp}
\vspace{-0.3cm}
\end{figure*}
\section{Experiments \& Results}
\subsection{Dataset}
We simulated training and test sets using three human-torso geometry models. Spatiotemporal TMP sequences were generated using the Aliev-Panfilov (AP) model \cite{panfilov96}, and projected to the body-surface potential data with 40dB SNR noises. %To consider generalization ability, we 
%Propagation of TMP in the heart is complex spatiotemporal pattern affected by numerous biophysical parameters. 
Two parameters were varied when simulating the TMP data : the origin of excitation and abnormal tissue properties representing myocardial scar. 
%We model the presence of scar at certain location of the heart with tissue properties. 

The training set was randomly selected 
with regard to these two parameters. 
%For the training set, we randomly picked locations on the heart as origin of excitation and randmly selected regions in the heart with different size as the scar region with different tissue properties. 
To test generalization ability, 
%To better access the inverse reconstruction and analyse generalization behavior of different methods, 
test data were selected 
with different origins of excitation and 
shape/location of abnormal tissues 
%scar shape and location 
than those used in training. 
%For each training set, 
In particular, we prepared test datasets of four types: 1) Scar: Low, Exc: Low, 2) Scar: Low, Exc: High, 3) Scar:High, Exc: Low, and 4) Scar: High, Exc: High, where Scar/Exc indicates the parameter being varied and High/Low denotes the 
level of difference from the training data.

%High, Exc:Low means test dataset contains TMP simulated with the location of scar not so different from that on training set but the origin of excitation point at very different location than that on training set. We prepared test dataset in this way for all three geometries.

\subsection{Implementation Details}
For all five models being compared (svs stochastic/deterministic, svs-L stochastic/ deterministic, and sss stochastic), we used ReLU activation functions in both the encoder and decoder, ADAM optimizer \cite{kingma2014adam}, and a learning rate of $10^{-3}$. 
Each neural network was trained on approximately 2500 TMP simulations on each geometry. In addition to the five neural networks, we included a classic TMP inverse reconstruction method (Greensite) designed to incorporate temporal information \cite{greensite98}. On each geometry, a random set of 100 test cases was selected from the test set and the process was repeated 120 times. We report the average and standard deviation of the results across all three geometry models.

\begin{figure*}[h!]
\centering
\includegraphics[width=0.8\linewidth]{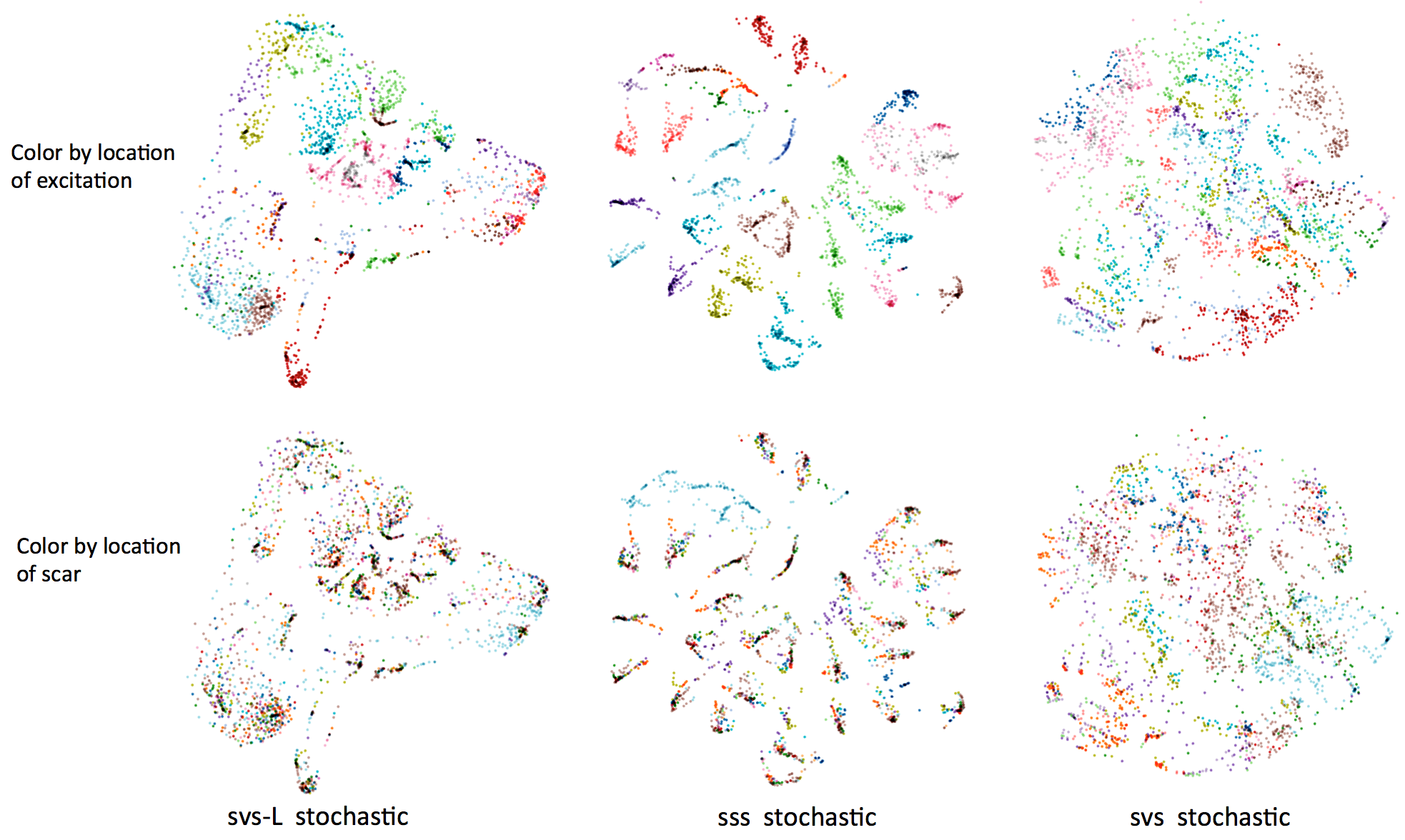}
\caption{\small{Visualization of latent point cloud corresponding to whole dataset: training and validation dataset.}}
\label{latent}
\vspace{-0.2cm}
\end{figure*}

\begin{figure*}[t!]
\centering
\includegraphics[width=0.8\linewidth]{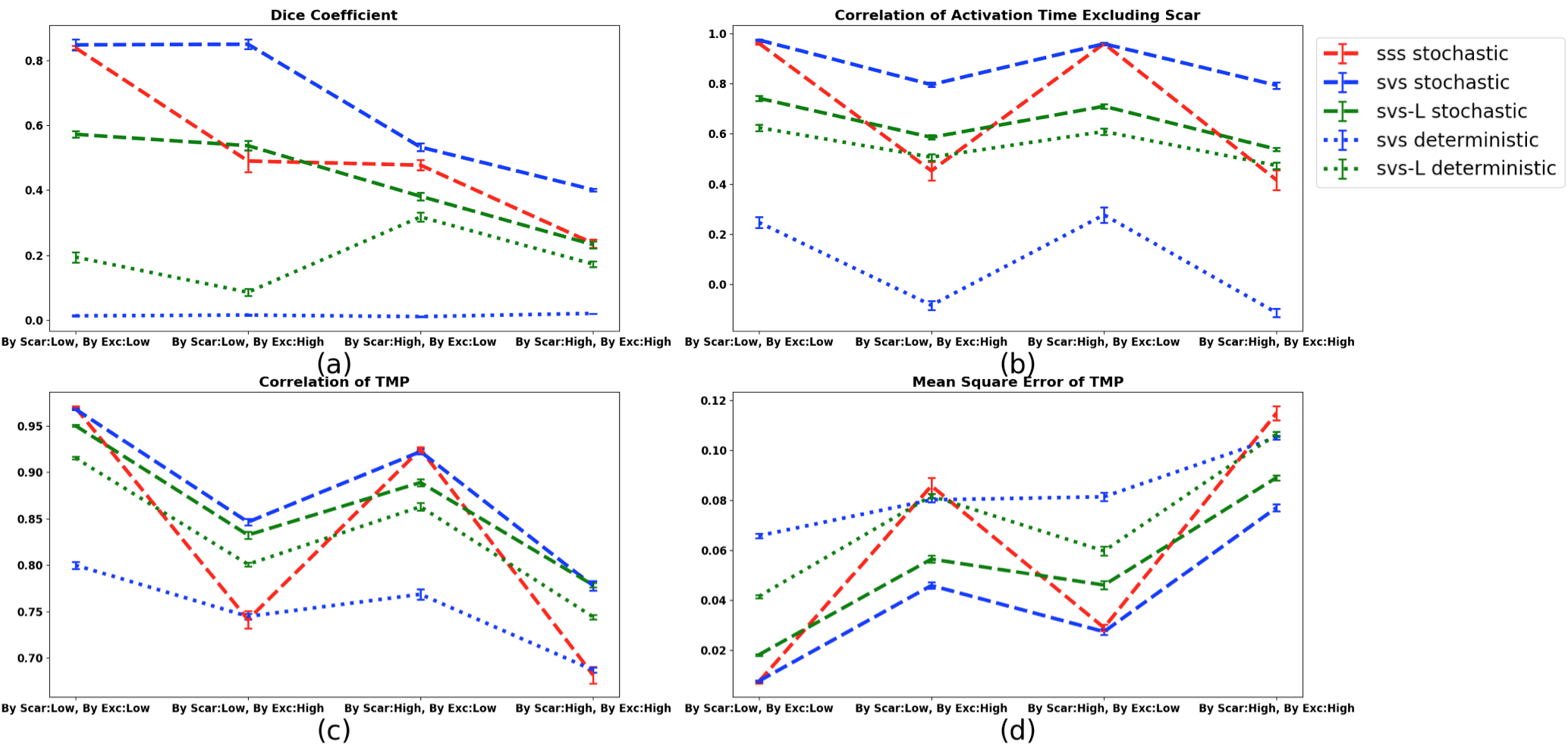}
\vspace{-0.5cm}
\caption{\small{Comparison of reconstruction ability of different architecture at the presence of different types of unseen test data.}}
\label{metrics_plot}
\end{figure*}

\begin{figure*}[tb!]
\centering
\includegraphics[width=0.75\linewidth]{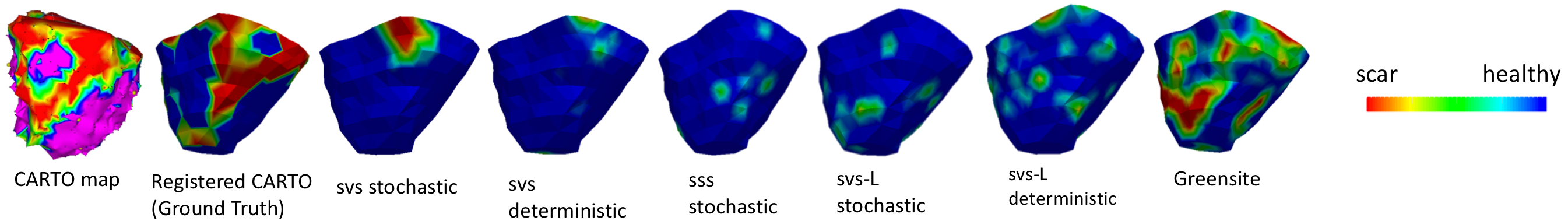}
\vspace{-0.3cm}
\caption{\small{Comparison of scar region identified by different architectures and Greensite method with reference to \textit{in vivo} CARTO map.}}
\label{realdata}
\vspace{-0.2cm}
\end{figure*}

\subsection{Results}
\begin{table}[tp]
\begin{center}

\caption{Metric for each method}
\begin{tabu} to 0.48\textwidth { | p{1.7cm}|| X[c] |X[c] | X[c] |  X[r] | }
 \hline
  Metric & MSE & TMP Corr. & AT Corr. & Dice Coeff.  \\  
 \hline\hline
 
 \begin{tabular}{@{}c@{}}svs \\ stochastic\end{tabular} & $\mathbf{0.037} \pm \mathbf{0.021}$ & $\mathbf{0.885} \pm \mathbf{0.061}$ & $\mathbf{0.885} \pm \mathbf{0.072}$& $\mathbf{0.645} \pm \mathbf{0.181}$ \\
 \hline
 \begin{tabular}{@{}c@{}}svs \\ deterministic\end{tabular}& $0.075 \pm 0.013$ & $0.77 \pm 0.038$ & $0.12 \pm 0.13$& $0.01 \pm 0.006$\\ 
 \hline
 \begin{tabular}{@{}c@{}}sss \\ stochastic\end{tabular} & $0.052 \pm 0.035$ & $0.847 \pm 0.09$ & $0.731 \pm 0.21$& $0.531 \pm 0.198$ \\ 
 \hline
 \begin{tabular}{@{}c@{}}svs-L \\ stochastic\end{tabular}& $0.068 \pm 0.023$ & $0.838 \pm 0.053$ & $0.601 \pm 0.074$& $0.28 \pm 0.154$\\
 \hline
 \begin{tabular}{@{}c@{}}svs-L \\ deterministic\end{tabular} & $0.067 \pm 0.02$ & $0.84 \pm 0.053$ & $0.57 \pm 0.052$& $0.165 \pm 0.092$ \\ 
 \hline
 \begin{tabular}{@{}c@{}}Greensite\end{tabular} & -- & -- & $0.514 \pm 0.006$& $0.138 \pm 0.005$ \\ 
 \hline
\end{tabu}
\end{center}
\vspace{-0.5cm}
\end{table}
The reconstruction accuracy was measured with four metrics: 1) mean square error (MSE) of the TMP sequence, 2) correlation of the TMP sequence, 3) correlation of TMP-derived activation time (AT), excluding late activation due to abnormal scar tissue to focus the measure on the accuracy related to excitation points, and 4) dice coefficients of the abnormal scar tissue identified from the TMP sequence. 
%Dice coefficient measures the accuracy with which scar region is identified while correlation coefficient excluding scar accuracy with which propagation sequence has been identified. The correlation coefficient of activation time, mean square error and correlation of TMP measure both the ability.

Table 1 summarizes the results from the three geometry models on all datasets. As shown, accuracy of the svs stochastic architecture was significantly higher than other architectures in all metrics. Similarly, the stochastic version of each architecture was more accurate than its deterministic counterpart. 
Most of the networks delivered a higher accuracy than the 
classic Greensite method (which does not preserve TMP signal shape and thus its MSE and correlation of TMP was not reported). %Because of incorporation of temporal information, inverse reconstructions have high correlation of AT but low dice coefficient. 
These observations are reflected in the example of reconstructed TMP sequences in Fig.~\ref{tmp}. 
%It compares TMP propagation over heart reconstructed from same ECG using different architecture. The propagation sequence is closer to ground truth in case of TMP reconstruction with svs stochastic architecture. 

%To better understand how the presented method performs better than others, we perform two more analysis.

\subsubsection{Analysis of latent representation:} To understand the difference in the latent representations obtained among different architectures, we computed latent projections of the training and validation sets for each method, and visualized the point cloud in the latent space with t-SNE \cite{maaten2008visualizing} as shown in Fig.\ref{latent}. Each data point has an excitation label and scar label corresponding to their locations in the heart.  
%based on the 17-segment definitions by the American heart association \cite{}. 
The first row shows latent points colored by the excitation label, suggesting that all three stochastic models were able to cluster data points in the latent space according to the origin of excitation. The second row shows latent points colored by scar label. In this case, only the latent cloud from the svs stochastic model was clustered. This suggests that the latent representation of the presented svs stochastic model considered information of both the scar location and origin of excitation, while the other models were more focused on the excitation point. 
%We may explain this behavior noting that both the methods svs-L and sss were designed with an emphasis on temporal consistency. Since temporal consistency is heavily affected by the origin of excitation, these methods tend to represent TMP sequence by emphasizing more on origin of excitation. svs-stochastic method, however, compresses both the information into one representation.
%\vspace{-0.2cm}
\subsubsection{Analysis of performance under different test conditions:} 

Figure \ref{metrics_plot} compares the performance of different methods in the four aforementioned types of test dataset, considering two levels of differences in each of the two parameters (excitation points and scar tissue) relative to the training data. 

When only the level of unseen excitation points increases in the new data (see Fig. \ref{metrics_plot}.a first two points on x axis), 
%with the scar region less different from the training data, 
%but the scar region remaining the same, 
the dice coefficient decreased in all architectures except for the svs stochastic model. This suggests that the representation of scar location in the svs stochastic model is more robust to errors in the excitation point.  
%while other methods depend more on the accuracy of origin of excitation. 
On the other hand, correlation of AT shows that when only the level of unseen scar tissues varies (see Fig. \ref{metrics_plot}.b first and third points on x axis), the performance of all methods stay at a constant level showing that they all encode the origin of excitation. These two findings were also consistent with the visualization of latent cloud in Fig. \ref{latent}. 

%In terms of correlation coefficient (c), we note that the two deterministic methods and sss are highly sensitive to the level of unseen data in the excitation set. In comparison, for svs stochastic and svs-L stochastic models, the overall error level depends on the level of unseen-ness in both the excitation and scar settings. 
 
Comparing all the metrics, we observed that even though all the models were trained on scaled mean square loss, the stochastic models and svs stochastic performed better in identifying origin of excitation and region of scar. 
%In specific, while the difference in correlation is not so significant between svs and svs stochastic at the presence of unseen data, the difference is much higher for the dice coefficient and correlation of AT. 
This suggests that stochasticity combined with aggregation of sequence information helps capture global generative factors and  thus improves generalization ability. 
\vspace{-0.1cm}
\subsubsection{Performance on Real Data:}
%To assess the feasibility of presented method and compare performance difference between different architecture choices on a real clinical application, 
We present a real-data case study %was performed on real data from
on a patient who underwent catheter ablation due to scar-related ventricular tachycardia. First, the presented models were trained on data simulated on this patient as described earlier. Then, TMP was reconstructed from the real ECG data using the trained networks, from which the scar region was delineated based on TMP duration and compared with low-voltage regions from \textit{in-vivo} mapping data. As shown in Fig. \ref{realdata}, the identified region of scar from the svs stochastic model is the closest to the \textit{in-vivo} data, which is consistent with simulated results. 
\vspace{-0.1cm}

%On generalization: How the performance is similar at the seen level and the difference in performance keeps appearing as we increase the level of unseen-ness.

%On the properties of general feature: The MSE might be similar, but looking at other metrics, we see that the reconstruction is better at capturing other essential patterns not captured by just measuring mSE.

%Latent feature analysis: What is the difference due stochasticity or compression? What is its advantage in terms of reconstruction? in Exc point? In scar? Some graphs? Some intuitions?
\section{Conclusion}
To our knowledge, this is the first work connecting VAE type of stochastic regularization with the generalization ability of a neural network. We have shown both theoretically and experimentally on inverse TMP reconstruction that the stochasticity and global aggregation of temporal information indeed improves inverse reconstruction. 
%We also found some interesting insights into the nature of latent representation and its effect on inverse reconstructions. 
Future works will extend these analyses on a wider variety of image and signal reconstruction problems over time sequences. 
%We intend to confirm these theories on other problems in the future.
%\clearpage
\bibliographystyle{aaai}
\bibliography{bibli1}

\begin{thebibliography}{}

\bibitem[\protect\citeauthoryear{Aliev and Panfilov}{1996}]{panfilov96}
Aliev, R.~R., and Panfilov, A.~V.
\newblock 1996.
\newblock A simple two-variable model of cardiac excitation.
\newblock {\em Chaos, Solitons \& Fractals} 7(3):293--301.

\bibitem[\protect\citeauthoryear{Bahdanau, Cho, and Bengio}{2014}]{bahdanau14}
Bahdanau, D.; Cho, K.; and Bengio, Y.
\newblock 2014.
\newblock Neural machine translation by jointly learning to align and
  translate.
\newblock {\em arXiv preprint arXiv:1409.0473}.

\bibitem[\protect\citeauthoryear{Bowman \bgroup et al\mbox.\egroup
  }{2015}]{bowman15}
Bowman, S.~R.; Vilnis, L.; Vinyals, O.; Dai, A.~M.; Jozefowicz, R.; and Bengio,
  S.
\newblock 2015.
\newblock Generating sentences from a continuous space.
\newblock {\em arXiv preprint arXiv:1511.06349}.

\bibitem[\protect\citeauthoryear{Chen \bgroup et al\mbox.\egroup
  }{2017}]{chen2017low}
Chen, H.; Zhang, Y.; Zhang, W.; Liao, P.; Li, K.; Zhou, J.; and Wang, G.
\newblock 2017.
\newblock Low-dose ct via convolutional neural network.
\newblock {\em Biomedical optics express} 8(2):679--694.

\bibitem[\protect\citeauthoryear{Fischer \bgroup et al\mbox.\egroup
  }{2015}]{fischer2015flownet}
Fischer, P.; Dosovitskiy, A.; Ilg, E.; H{\"a}usser, P.; Haz{\i}rba{\c{s}}, C.;
  Golkov, V.; Van~der Smagt, P.; Cremers, D.; and Brox, T.
\newblock 2015.
\newblock Flownet: Learning optical flow with convolutional networks.
\newblock {\em arXiv preprint arXiv:1504.06852}.

\bibitem[\protect\citeauthoryear{Graves}{2013}]{graves2013generating}
Graves, A.
\newblock 2013.
\newblock Generating sequences with recurrent neural networks.
\newblock {\em arXiv preprint arXiv:1308.0850}.

\bibitem[\protect\citeauthoryear{Greensite and Huiskamp}{1998}]{greensite98}
Greensite, F., and Huiskamp, G.
\newblock 1998.
\newblock An improved method for estimating epicardial potentials from the body
  surface.
\newblock {\em IEEE TBME} 45(1):98--104.

\bibitem[\protect\citeauthoryear{Hardy}{1906}]{hardy}
Hardy, G.~H.
\newblock 1906.
\newblock On double fourier series and especially those which represent the
  double zeta-function with real and incommensurable parameters.
\newblock {\em Quart. J. Math} 37(5).

\bibitem[\protect\citeauthoryear{Jin \bgroup et al\mbox.\egroup }{2017}]{jin17}
Jin, K.~H.; McCann, M.~T.; Froustey, E.; and Unser, M.
\newblock 2017.
\newblock Deep convolutional neural network for inverse problems in imaging.
\newblock {\em IEEE Transactions on Image Processing} 26(9):4509--4522.

\bibitem[\protect\citeauthoryear{Kalchbrenner and
  Blunsom}{2013}]{kalchbrenner2013recurrent}
Kalchbrenner, N., and Blunsom, P.
\newblock 2013.
\newblock Recurrent continuous translation models.
\newblock In {\em Proceedings of the 2013 Conference on Empirical Methods in
  Natural Language Processing},  1700--1709.

\bibitem[\protect\citeauthoryear{Kawaguchi and
  Bengio}{2018}]{kawaguchi18analytical}
Kawaguchi, K., and Bengio, Y.
\newblock 2018.
\newblock Generalization in machine learning via analytical learning theory.
\newblock {\em arXiv preprint arXiv:1802.07426}.

\bibitem[\protect\citeauthoryear{Kingma and Ba}{2014}]{kingma2014adam}
Kingma, D.~P., and Ba, J.
\newblock 2014.
\newblock Adam: A method for stochastic optimization.
\newblock {\em arXiv preprint arXiv:1412.6980}.

\bibitem[\protect\citeauthoryear{Kingma and Welling}{2013}]{kingma13}
Kingma, D.~P., and Welling, M.
\newblock 2013.
\newblock Auto-encoding variational bayes.
\newblock {\em arXiv preprint arXiv:1312.6114}.

\bibitem[\protect\citeauthoryear{Lipton \bgroup et al\mbox.\egroup
  }{2015}]{lipton15}
Lipton, Z.~C.; Kale, D.~C.; Elkan, C.; and Wetzel, R.
\newblock 2015.
\newblock Learning to diagnose with lstm recurrent neural networks.
\newblock {\em arXiv preprint arXiv:1511.03677}.

\bibitem[\protect\citeauthoryear{Lucas \bgroup et al\mbox.\egroup
  }{2018}]{lucas18}
Lucas, A.; Iliadis, M.; Molina, R.; and Katsaggelos, A.~K.
\newblock 2018.
\newblock Using deep neural networks for inverse problems in imaging: beyond
  analytical methods.
\newblock {\em IEEE Signal Processing Magazine} 35(1):20--36.

\bibitem[\protect\citeauthoryear{Luong, Pham, and Manning}{2015}]{luong15}
Luong, M.-T.; Pham, H.; and Manning, C.~D.
\newblock 2015.
\newblock Effective approaches to attention-based neural machine translation.
\newblock {\em arXiv preprint arXiv:1508.04025}.

\bibitem[\protect\citeauthoryear{Maaten and
  Hinton}{2008}]{maaten2008visualizing}
Maaten, L. v.~d., and Hinton, G.
\newblock 2008.
\newblock Visualizing data using t-sne.
\newblock {\em JMLR} 9(Nov):2579--2605.

\bibitem[\protect\citeauthoryear{MacLeod \bgroup et al\mbox.\egroup
  }{1995}]{macleod95}
MacLeod, R.~S.; Gardner, M.; Miller, R.~M.; and HOR{\'A}C̆UEK, B.~M.
\newblock 1995.
\newblock Application of an electrocardiographic inverse solution to localize
  ischemia during coronary angioplasty.
\newblock {\em Journal of cardiovascular electrophysiology} 6(1):2--18.

\bibitem[\protect\citeauthoryear{Mao, Shen, and Yang}{2016}]{mao16}
Mao, X.; Shen, C.; and Yang, Y.-B.
\newblock 2016.
\newblock Image restoration using very deep convolutional encoder-decoder
  networks with symmetric skip connections.
\newblock In {\em Advances in neural information processing systems},
  2802--2810.

\bibitem[\protect\citeauthoryear{Pathak \bgroup et al\mbox.\egroup
  }{2016}]{pathak2016context}
Pathak, D.; Krahenbuhl, P.; Donahue, J.; Darrell, T.; and Efros, A.~A.
\newblock 2016.
\newblock Context encoders: Feature learning by inpainting.
\newblock In {\em Proceedings of the IEEE Conference on Computer Vision and
  Pattern Recognition},  2536--2544.

\bibitem[\protect\citeauthoryear{Plonsey}{1969}]{plonsey1969bioelectric}
Plonsey, R.
\newblock 1969.
\newblock Bioelectric phenomena.

\bibitem[\protect\citeauthoryear{Ramanathan \bgroup et al\mbox.\egroup
  }{2004}]{ramanathan2004}
Ramanathan, C.; Ghanem, R.~N.; Jia, P.; Ryu, K.; and Rudy, Y.
\newblock 2004.
\newblock Noninvasive electrocardiographic imaging for cardiac
  electrophysiology and arrhythmia.
\newblock {\em Nature medicine} 10(4):422.

\bibitem[\protect\citeauthoryear{Sutskever, Vinyals, and
  Le}{2014}]{sutskever14}
Sutskever, I.; Vinyals, O.; and Le, Q.~V.
\newblock 2014.
\newblock Sequence to sequence learning with neural networks.
\newblock In {\em Advances in neural information processing systems},
  3104--3112.

\bibitem[\protect\citeauthoryear{Wang \bgroup et al\mbox.\egroup
  }{2010}]{wang10}
Wang, L.; Zhang, H.; Wong, K.~C.; Liu, H.; and Shi, P.
\newblock 2010.
\newblock Physiological-model-constrained noninvasive reconstruction of
  volumetric myocardial transmembrane potentials.
\newblock {\em IEEE Transactions on Biomedical Engineering} 57(2):296--315.

\bibitem[\protect\citeauthoryear{Wang \bgroup et al\mbox.\egroup
  }{2013}]{wang13}
Wang, L.; Dawoud, F.; Yeung, S.-K.; Shi, P.; Wong, K.~L.; Liu, H.; and Lardo,
  A.~C.
\newblock 2013.
\newblock Transmural imaging of ventricular action potentials and
  post-infarction scars in swine hearts.
\newblock {\em IEEE TMI} 32(4):731--747.

\bibitem[\protect\citeauthoryear{Wang \bgroup et al\mbox.\egroup
  }{2015}]{wang2015deep}
Wang, Z.; Liu, D.; Yang, J.; Han, W.; and Huang, T.
\newblock 2015.
\newblock Deep networks for image super-resolution with sparse prior.
\newblock In {\em Proceedings of the IEEE International Conference on Computer
  Vision},  370--378.

\bibitem[\protect\citeauthoryear{Wang \bgroup et al\mbox.\egroup
  }{2016}]{wang2016accelerating}
Wang, S.; Su, Z.; Ying, L.; Peng, X.; Zhu, S.; Liang, F.; Feng, D.; and Liang,
  D.
\newblock 2016.
\newblock Accelerating magnetic resonance imaging via deep learning.
\newblock In {\em Biomedical Imaging (ISBI), 2016 IEEE 13th International
  Symposium on},  514--517.
\newblock IEEE.

\bibitem[\protect\citeauthoryear{Wang \bgroup et al\mbox.\egroup
  }{2018}]{Wang_Europace18}
Wang, L.; Gharbia, O.~A.; Horacek, B.~M.; Nazarian, S.; ; and Sapp, J.~L.
\newblock 2018.
\newblock Noninvasive epicardial and endocardial electrocardiographic imaging
  of scar-related ventricular tachycardia.
\newblock {\em EP Europace}  under review after major revision.

\bibitem[\protect\citeauthoryear{Werbos}{1990}]{werbos1990backpropagation}
Werbos, P.~J.
\newblock 1990.
\newblock Backpropagation through time: what it does and how to do it.
\newblock {\em Proceedings of the IEEE} 78(10):1550--1560.

\bibitem[\protect\citeauthoryear{Yao \bgroup et al\mbox.\egroup
  }{2017}]{yao2017dr2}
Yao, H.; Dai, F.; Zhang, D.; Ma, Y.; Zhang, S.; Zhang, Y.; and Tian, Q.
\newblock 2017.
\newblock Dr2-net: Deep residual reconstruction network for image compressive
  sensing.
\newblock {\em arXiv preprint arXiv:1702.05743}.

\bibitem[\protect\citeauthoryear{Yue-Hei~Ng \bgroup et al\mbox.\egroup
  }{2015}]{yue15}
Yue-Hei~Ng, J.; Hausknecht, M.; Vijayanarasimhan, S.; Vinyals, O.; Monga, R.;
  and Toderici, G.
\newblock 2015.
\newblock Beyond short snippets: Deep networks for video classification.
\newblock In {\em Proceedings of the IEEE conference on computer vision and
  pattern recognition},  4694--4702.

\bibitem[\protect\citeauthoryear{Zhu \bgroup et al\mbox.\egroup }{2018}]{zhu18}
Zhu, B.; Liu, J.~Z.; Cauley, S.~F.; Rosen, B.~R.; and Rosen, M.~S.
\newblock 2018.
\newblock Image reconstruction by domain-transform manifold learning.
\newblock {\em Nature} 555(7697):487.

\end{thebibliography}

\end{document}